\documentclass[11pt]{article}

\usepackage[final]{acl}

\usepackage{times}
\usepackage{latexsym}

\usepackage{algorithm}
\usepackage{algpseudocode}
\usepackage{amsmath, amssymb, amsfonts}
\usepackage{textcomp}
\usepackage{xcolor}
\usepackage{booktabs} 
\usepackage{multirow}
\usepackage{hyperref} 
\usepackage{enumitem} 
\usepackage{dsfont}

\usepackage[T1]{fontenc}

\usepackage[utf8]{inputenc}

\usepackage{microtype}

\usepackage{inconsolata}

\usepackage{graphicx}

\title{Not All Memories Are Equal: Hierarchical Collaborative Memory for Validity-Aware Retrieval in LLM Agents}

\author{
 \textbf{Yufei Shi\textsuperscript{1}},
 \textbf{Rujing Yao\textsuperscript{1}},
 \textbf{Ang Li\textsuperscript{2}},
 \textbf{Yang Wu\textsuperscript{4}},\\
 \textbf{Zhuoren Jiang\textsuperscript{3}\textsuperscript{*}},
 \textbf{Xiaozhong Liu\textsuperscript{4}\textsuperscript{*}}
\\
 \textsuperscript{1}Nanyang Technological University, Singapore \\
 \textsuperscript{2}University of Macau, China \\
 \textsuperscript{3}Zhejiang University, China \\
 \textsuperscript{4}Worcester Polytechnic Institute, USA
\\
\small\texttt{shiy0065@e.ntu.edu.sg, rujing.yao@ntu.edu.sg, leeyonli@um.edu.mo,}\\
\small\texttt{ywu19@wpi.edu, jiangzhuoren@zju.edu.cn, xliu14@wpi.edu}}

\begin{document}
\maketitle
\begin{abstract}
In team collaboration scenarios, memory is heterogeneous and continually evolving. Team memories capture collective decisions, protocols, and current consensus, while individual memories preserve member-specific observations, execution traces, and intermediate progress. Existing memory-augmented systems typically retrieve from all stored memories as a flat pool, ranking them by semantic relevance, importance, or recency without modeling hierarchical structure or evolving validity. As a result, they often surface semantically relevant but outdated or conflicting memories, especially individual memories that no longer align with current team consensus, instead of prioritizing currently valid memories. This is particularly problematic when collaborative LLM agents answer user questions, since their responses should be grounded in valid memories. We propose HiCoMER, a framework for hierarchical collaborative memory management and validity-aware retrieval in LLM agents. HiCoMER first maintains the validity of team and individual memories and then retrieves memories that remain valid, rather than retrieving directly from all stored memories. It consists of three components: a Hierarchical Memory Conflict Updater, a Validity-Aware Memory Retriever, and a Memory-Grounded Answer Generator. To evaluate HiCoMER, we construct two new datasets for memory-grounded question answering in collaborative settings. Experiments on both datasets show that HiCoMER consistently outperforms strong baselines by reducing outdated retrieval, preserving current team consensus, and improving downstream QA quality.
\end{abstract}

\renewcommand{\thefootnote}{\fnsymbol{footnote}}
\footnotetext[1]{Corresponding authors.}
\renewcommand{\thefootnote}{\arabic{footnote}}

\section{Introduction}
\label{sec:introduction}

Large language model (LLM) agents are increasingly moving beyond single-turn assistance toward long-horizon collaboration in shared environments~\cite{boiko2023autonomous,gao2024biomedical}. In these settings, memory is heterogeneous and hierarchical: team memories record collective decisions and protocols, while individual memories capture member-specific observations and execution traces~\cite{shuster2022blenderbot}.
\begin{figure}[t]
    \centering
    \includegraphics[width=1\linewidth]{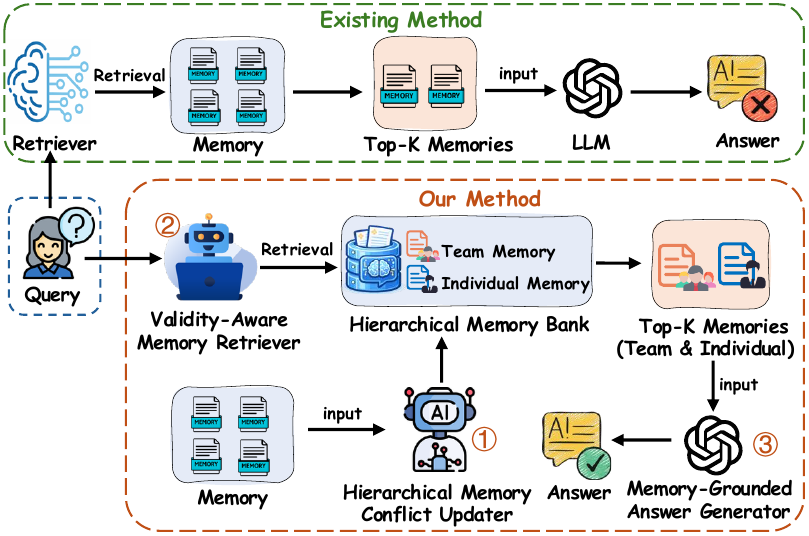}
    \caption{Illustration comparing existing methods and our proposed HiCoMER.}
    \label{fig:fig1}
\end{figure}
Agents may retrieve semantically relevant but invalid memories, overlooking current team consensus. For example, when asked ``How should we synthesize Compound A?'', an agent may select a past individual experiment log while ignoring a team-level update banning Compound A due to toxicity. Existing memory-augmented systems~\cite{lewis2020retrieval,packer2023memgpt} retrieve from a flat pool based on semantic similarity or recency, failing to capture hierarchical conflicts or evolving validity, as shown in Figure~\ref{fig:fig1}.

Moreover, collaborative memories are often produced in temporally aligned groups. A single meeting or experiment cycle may generate both team-level decisions and multiple individual records, some of which may conflict with the current consensus. Memory management must therefore account for both within-group hierarchical conflicts and cross-time validity changes.

To address these challenges, we propose HiCoMER, a hierarchical collaborative memory framework. HiCoMER maintains memory validity and performs validity-aware retrieval through three components: a Hierarchical Memory Conflict Updater, a Validity-Aware Memory Retriever, and a Memory-Grounded Answer Generator. This ensures agents prioritize currently valid memories. We construct two new datasets for memory-grounded question answering in collaborative settings. Experiments show that HiCoMER consistently reduces outdated retrieval, preserves team consensus, and improves downstream QA quality.

Our contributions are as follows:
\begin{itemize}
    \item We formalize the problem of hierarchical memory conflict in collaborative LLM settings, where both team and individual memories may evolve over time, and semantically relevant memories can become invalid under the current consensus.
    \item We propose HiCoMER, a hierarchical collaborative memory framework that maintains memory validity and performs conflict-aware, validity-aware retrieval. HiCoMER consists of three components: a Hierarchical Memory Conflict Updater, a Validity-Aware Memory Retriever, and a Memory-Grounded Answer Generator, which together enable agents to prioritize currently valid memories.
    \item We construct two new datasets for memory-grounded question answering in collaborative environments. Experiments demonstrate that HiCoMER consistently improves retrieval safety, consensus preservation, and downstream QA performance over strong baselines.
\end{itemize}

\section{Related Work}
\vspace{-0.05in}
\subsection{Long-Term Memory Retrieval}
Retrieval-augmented generation (RAG) grounds LLM outputs on retrieved evidence and is widely used in knowledge-intensive NLP~\cite{lewis2020retrieval}. Many works improve the retrieval backbone, including dense retrievers for open-domain QA such as DPR~\cite{karpukhin2020dpr}, latent-retrieval pretraining like REALM~\cite{guu2020realm}, and large-scale retrieval-augmented pretraining as in RETRO~\cite{borgeaud2022retro}. On the reader side, fusion-based architectures such as FiD~\cite{izacard2021fid} and end-to-end retrieval-augmented models like Atlas~\cite{izacard2023atlas} enhance robustness and integration. Complementary IR advances include late interaction models like ColBERT~\cite{khattab2020colbert}, sparse expansion models like SPLADE~\cite{formal2021splade}, and query-side augmentation such as HyDE~\cite{gao2023hyde}. Self-reflective pipelines like Self-RAG critique retrieved evidence and generations to improve faithfulness~\cite{asai2024self}. Agent-oriented systems treat memory as a persistent, growing corpus with write/read policies, e.g., MemGPT~\cite{packer2023memgpt} and interactive generative agents~\cite{park2023generativeagents}. Despite these advances, most approaches optimize flat relevance signals and do not explicitly model hierarchical validity or authority, which is central to our setting.

\subsection{Hierarchical Consistency under Conflicts}

Conflict resolution is often studied through contradiction detection, entailment reasoning, and factual verification. NLI datasets like SNLI and MultiNLI provide supervision for entailment and contradiction~\cite{bowman2015snli,williams2018mnli}, and adversarial benchmarks such as ANLI stress-test robustness~\cite{nie2020anli}. Fact verification datasets like FEVER formalize consistency as verifying claims against evidence~\cite{thorne2018fever}. For generation, approaches include FactCC~\cite{kryscinski2020factcc}, QAGS-style QA checks~\cite{wang2020qags}, and broader evaluations like TruthfulQA~\cite{lin2022truthfulqa} and TRUE~\cite{honovich2022true}. Post-hoc detection and correction methods include SelfCheckGPT~\cite{manakul2023selfcheckgpt}, iterative retrieval-revision pipelines like RARR~\cite{gao2023rarr}, and fine-grained factual scoring such as FActScore~\cite{min2023factscore}. However, these works mostly resolve conflicts at generation or evaluation time, rather than as write-time operations on evolving memories. In collaborative environments, conflicts are frequently asymmetric: team decisions and SOP updates can invalidate many individual logs even if semantically similar to future queries. HiCoMER addresses this by organizing collaborative memories as time-aligned groups, performing trainable maintenance over hierarchical and temporal conflicts, and learning a validity-aware retrieval function over the maintained memory bank.

\section{Methodology}
\vspace{-0.05in}
\subsection{Framework Overview}

\label{sec:overview}
\begin{figure*}[ht]
\setlength{\belowcaptionskip}{-0cm}
    \centering
    \includegraphics[width=1\linewidth]{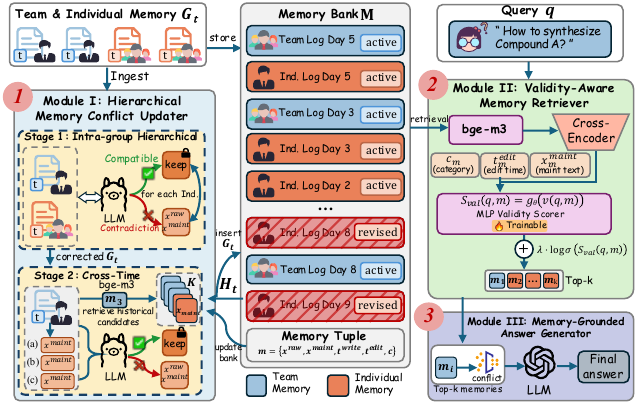}
    \caption{Overall framework of HiCoMER. The framework consists of three modules: Hierarchical Memory Conflict Updater, Validity-Aware Memory Retriever, and Memory-Grounded Answer Generator.}
    \label{fig:fig2}
\end{figure*}
HiCoMER is a collaborative memory framework for long-horizon LLM agents operating in team environments. In collaborative settings, memory is inherently heterogeneous: Team memories capture shared decisions, protocols, and current coordination state, while Individual memories record local observations, execution traces, and intermediate progress. Memory retrieval should therefore not rely solely on semantic relevance, because semantically relevant memories may already be outdated or conflict with other memories under the current collaborative state. This issue becomes particularly severe when agents retrieve directly from raw stored memories without explicitly modeling the distinct functions of Team and Individual memories or maintaining the memory bank to resolve outdated and conflicting memories. To address this problem, HiCoMER explicitly models collaborative memory at both the Team and Individual levels and maintains the memory bank to preserve a valid collaborative memory state for downstream question answering.

As illustrated in Figure~\ref{fig:fig2}, HiCoMER consists of three modules. The first module, Hierarchical Memory Conflict Updater, maintains the memory bank under two types of conflict: same-time hierarchical conflict between Team and Individual memories in the current group, and cross-time conflict between the current maintained Team/Individual memories and previously stored memories. The second module, Validity-Aware Memory Retriever, ranks candidate memories by combining semantic relevance with maintained memory-state signals, so that retrieval is guided not only by textual relevance but also by whether a memory remains valid under the current collaborative state. The third module, Memory-Grounded Answer Generator, produces the final response based on the retrieved memories after a lightweight conflict-resolution step.

Unlike conventional memory-augmented systems that retrieve directly from raw stored memories, HiCoMER explicitly models Team and Individual memories and maintains their evolving validity over time. This design allows the framework to reduce the retrieval of conflicting or outdated memories and ultimately generate answers grounded in currently valid memories rather than raw historical accumulation.

\subsection{Hierarchical Memory Conflict Updater}
\label{sec:conflictupdater}

This module maintains the collaborative memory bank by resolving outdated and conflicting memories under the hierarchical structure of Team and Individual memories. Its goal is to preserve a valid collaborative memory state for downstream retrieval and answer generation.

At each time step $t$, HiCoMER receives a time-aligned memory group
\begin{equation}
G_t = M_t^{\mathrm{Team}} \cup M_t^{\mathrm{Individual}},
\end{equation}
where $M_t^{\mathrm{Team}}$ contains collective records such as meeting resolutions, protocol updates, and supervisor instructions, while $M_t^{\mathrm{Individual}}$ contains individual records such as execution logs, observations, failed attempts, and intermediate progress. During maintenance, each memory is associated with both its original content and its maintained content, so that HiCoMER preserves the historical record while updating the currently valid memory view used by downstream modules.

HiCoMER first identifies a bounded set of relevant historical memories, because the current group is unlikely to affect the entire historical memory bank and exhaustively comparing against all past memories would be computationally inefficient. Specifically, it constructs a textual query $q_t^{\mathrm{hist}}$ from the current Team and Individual memories and retrieves the Top-$K$ most relevant memories from the previously stored bank:
\begin{equation}
H_t = \operatorname{TopK}_{m \in \mathcal{M}_{<t}} \; s_{\phi}\!\left(q_t^{\mathrm{hist}}, m\right),
\end{equation}
where $\mathcal{M}_{<t}$ denotes all Team and Individual memories stored before time $t$, $q_t^{\mathrm{hist}}$ is constructed from the current group $G_t$, and $s_{\phi}(\cdot,\cdot)$ is a dense semantic retrieval model that computes the relevance between the current-group query and each historical memory. $H_t$ contains the Top-$K$ historical Team/Individual memories with the highest retrieval scores.

Given the current group $G_t$ and the retrieved historical candidates $H_t$, the updater performs maintenance at two levels. The first level updates the current group itself. Within $G_t$, Team and Individual memories may express inconsistent states. HiCoMER therefore performs an intra-group hierarchical update that revises the current Individual memories under the coordination signal provided by the current Team memories, yielding a maintained current-group state:
\begin{equation}
\widetilde{G}_t
=
M_t^{\mathrm{Team}}
\cup
\widetilde{M}_t^{\mathrm{Individual}},
\end{equation}
where $\widetilde{G}_t$ denotes the maintained version of the current group after conflict resolution. In this stage, the Team memories are treated as the authoritative coordination state at time $t$, and the updater revises Individual memories when they are incompatible with that state.

The second level updates the retrieved historical candidates. HiCoMER uses the maintained current-group state $\widetilde{G}_t$, which consists of the current Team memories and the updated current Individual memories, to perform an inter-group hierarchical update over the historical candidates in $H_t$. This step revises previously stored Team and Individual memories whose maintained content has become outdated or conflicting under the current state.

We instantiate the updater with a lightweight instruction-tuned LLM and train it to generate structured maintenance decisions rather than free-form outputs. For each target memory, the model predicts three elements: a relation label, an action label, and a revised maintained text when revision is required. The relation label is drawn from $\{\mathrm{compatible}, \mathrm{neutral}, \mathrm{contradiction}\}$, and the action label is drawn from $\{\mathrm{keep}, \mathrm{revise}\}$. When the predicted action is ``revise", HiCoMER updates the maintained content.

We train the updater in two stages. We first perform supervised fine-tuning on structured maintenance supervision, so that the model learns to produce the desired decision format and content. We then further optimize it using Group Relative Policy Optimization (GRPO)~\cite{shao2024deepseekmath}. Given an input context $u$, the policy samples a set of candidate structured outputs
\begin{equation}
\mathcal{Y}(u)=\{\tilde{y}_1,\ldots,\tilde{y}_G\},
\end{equation}
which are scored by a deterministic reward
\begin{equation}
R(\tilde{y};u)=\lambda_1 R_{\mathrm{cons}}+\lambda_2 R_{\mathrm{rev}}+\lambda_3 R_{\mathrm{fmt}},
\end{equation}
where $R_{\mathrm{cons}}$ measures consistency with the gold maintained state, $R_{\mathrm{rev}}$ encourages necessary but minimal revision, and $R_{\mathrm{fmt}}$ enforces schema-valid output. In this way, GRPO encourages the updater to generate maintenance decisions that are accurate, concise, and structurally valid.

\subsection{Validity-Aware Memory Retriever}
\label{sec:retriever}

This module retrieves memories that are not only semantically relevant to the query but also valid under the current maintained collaborative state. Its goal is to reduce the retrieval of outdated or conflicting memories while preserving evidence that remains useful for downstream answer generation.

Given a user query $q$ and the maintained memory bank $\mathcal{M}_{\le t}$ after processing groups up to time $t$, HiCoMER first retrieves a high-recall candidate set
\begin{equation}
R_t(q)=\operatorname{TopN}_{m \in \mathcal{M}_{\le t}} \; r_{\psi}(q,m),
\end{equation}
where $r_{\psi}(\cdot,\cdot)$ is a dense retrieval model that measures the semantic relevance between the query and each maintained memory, and $R_t(q)$ contains the top-$N$ candidate memories returned by the first-stage retriever.

HiCoMER then reranks each candidate memory $m \in R_t(q)$ using two complementary signals.

The first is a semantic relevance score
\begin{equation}
S_{\mathrm{sem}}(q,m)=f_{\omega}(q,m),
\end{equation}
where $f_{\omega}(\cdot,\cdot)$ is a cross-encoder reranker that jointly encodes the query and the candidate memory for fine-grained relevance scoring.

The second is a validity-aware score that reflects whether a candidate memory remains usable under the current maintained state. For each candidate memory, HiCoMER considers its maintained content, its source type (Team or Individual), and its most recent maintenance time. Let $h_q$ and $h_m^{\mathrm{maint}}$ denote the dense representations of the query and the maintained content of memory $m$, respectively. HiCoMER then constructs a feature vector
\begin{equation}
v(q,m)=
[h_m^{\mathrm{maint}} ; h_q \odot h_m^{\mathrm{maint}} ; e(c_m) ; \tau_m],
\end{equation}
where $e(c_m)$ is a trainable embedding of the memory type $c_m \in \{\mathrm{Team},\mathrm{Individual}\}$, and $\tau_m$ is a normalized timestamp feature derived from the latest maintenance time of $m$. A lightweight MLP produces the validity-aware score
\begin{equation}
S_{\mathrm{val}}(q,m)=g_{\theta}(v(q,m)).
\end{equation}

HiCoMER combines these two signals into a unified retrieval score:
\begin{equation}
S_{\mathrm{final}}(q,m)
=
S_{\mathrm{sem}}(q,m)
+
\lambda \cdot \log \sigma\!\left(S_{\mathrm{val}}(q,m)\right),
\end{equation}
where $\lambda$ controls the contribution of the validity-aware component. The final retrieved evidence is obtained by ranking candidate memories in $R_t(q)$ according to $S_{\mathrm{final}}(q,m)$.

We train the retriever after obtaining maintained memory states from Section~\ref{sec:conflictupdater}. For each query $q$, we construct pairwise ranking instances consisting of a positive memory $m^{+}$ and a negative memory $m^{-}$, where $m^{+}$ is a gold supporting memory that remains valid under the maintained state, while $m^{-}$ is an outdated, conflicting, or semantically similar but non-valid competitor. The retriever is optimized with a pairwise ranking objective
\begin{equation}
\mathcal{L}_{\mathrm{retr}}
=
-\log
\left(
\frac{\exp(s^{+})}
{\exp(s^{+})+\exp(s^{-})}
\right).
\end{equation}
so that the validity-aware scorer learns to rank currently valid memories above outdated or conflicting alternatives.

At inference time, HiCoMER first retrieves a candidate set using the dense retriever and then reranks the candidates using the unified score $S_{\mathrm{final}}(q,m)$. The resulting ranked memories are passed to the downstream answer generator.

\subsection{Memory-Grounded Answer Generator}
\label{sec:generator}

This module generates the final answer from the retrieved memories. Its goal is to produce responses grounded in the retrieved evidence while suppressing residual conflicts that may still remain among the top-ranked candidates.

Given a user query $q$ and the ranked memory set returned by the retriever, a lightweight conflict-resolution step is applied over the retrieved evidence. Although Module I and Module II already reduce outdated and conflicting memories, residual inconsistency may still remain when multiple memories are semantically relevant to the same query but reflect different collaborative states.

To address this issue, each retrieved memory is assigned a priority score based on its source type, maintained validity, and maintenance time:

\begin{equation}
\begin{split}
P(m) &= \alpha \cdot \mathbb{I}[c_m=\mathrm{Team}] \\
     &\quad + \beta \cdot \mathbb{I}[\mathrm{Valid}(m)] \\
     &\quad + \gamma \cdot \mathrm{NormTime}(t_m^{edit}),
\end{split}
\end{equation}
where $\alpha$, $\beta$, and $\gamma$ are positive weights, $c_m$ denotes the memory type, and $\mathrm{NormTime}(t_m^{edit})$ is the normalized maintenance-time score. If two retrieved memories have contradictory maintained contents, the memory with higher priority is retained. Detailed conflict-resolution rules are provided in Appendix~\ref{app:conflict_resolution}. The resulting evidence set is then passed to the answer generator.

The generator receives the query together with the resolved evidence set and produces the final response based solely on this evidence.

\section{Experiments}
\label{sec:experiments}

\subsection{Dataset}
\label{sec:dataset}
Existing datasets lack explicit hierarchical conflicts and temporally-aligned team versus individual memories, which are crucial for evaluating memory maintenance and validity-aware retrieval in collaborative long-horizon tasks. To address this gap, we construct two new datasets corresponding to two biomedical collaboration settings: Acute Lung Injury (ALI) and PROTAC platform (PROTAC). 

The ALI dataset focuses on sepsis-related therapeutic development, where conflicts are primarily driven by safety-sensitive protocol updates, treatment discontinuation, and endpoint revisions. The PROTAC dataset models linker-free PROTAC platform development under N-end rule constraints, with conflicts arising more often from target-strategy revisions, assay replacements, and platform-level design invalidations.

\begin{table}[t]
\caption{Detailed statistics.}
\label{tab:dataset_1}
\centering
\resizebox{\columnwidth}{!}{
\begin{tabular}{lccc}
\toprule
\textbf{Statistic} & \textbf{ALI} & \textbf{PROTAC} & \textbf{Total} \\
\midrule
Candidate papers & 30 & 30 & 60 \\
Retained papers & 27 & 26 & 53 \\
Valid groups & 908 & 873 & 1,781 \\
Memory documents & 8,917 & 8,801 & 17,718 \\
Injected conflicts & 317 & 304 & 621 \\
Same-time conflicts & 189 & 181 & 370 \\
Cross-time conflicts & 128 & 123 & 251 \\
Train papers & 19 & 18 & 37 \\
Validation papers & 3 & 3 & 6 \\
Test papers & 5 & 5 & 10 \\
\bottomrule
\end{tabular}
}
\end{table}

The detailed dataset statistics are summarized in Table~\ref{tab:dataset_1}. The full construction process is provided in Appendix~\ref{sec:appendix_dataset}, and illustrative examples are provided in Appendix~\ref{sec:appendix_illustrative_examples}.

\subsection{Baselines and Evaluation Metrics}
\label{sec:baselines_metrics}

We compare HiCoMER against a diverse set of baselines covering standard semantic retrieval pipelines such as flat RAG, hybrid RAG, and rerank RAG; recency-based memory control methods; agentic and hierarchical memory frameworks including MemGPT-style agents~\cite{packer2023memgpt} and G-Memory~\cite{zhang2025gmemory}; and reflection- or production-oriented systems that rely on generation-time reasoning without explicit memory-bank maintenance, such as Self-RAG~\cite{asai2024self} and Mem0~\cite{chhikara2025mem0}. These baselines are chosen to test whether hierarchical conflict resolution can be addressed by stronger semantic matching alone, by recency heuristics, by existing long-horizon memory mechanisms, or by generation-time reflection without explicit memory maintenance. We also include ablated variants of HiCoMER to isolate the contributions of the Hierarchical Memory Conflict Updater and the Validity-Aware Memory Retriever. For fairness, all methods are evaluated on the same grouped memory stream and query interface, and system-level baselines are adapted to the same retrieval settings. Full baseline definitions are provided in Appendix~\ref{sec:appendix_baselines_metrics}.

Performance is evaluated at three levels. For the Hierarchical Memory Conflict Updater, we measure structured maintenance quality using Conflict F1 and Maintenance Accuracy. For upstream maintained-state retrieval, we use Outdated Retrieval Rate at 5 (ORR@5) to quantify the reduction of outdated memory retrieval, Consensus Retention Rate at 5 (CRR@5) to evaluate preservation of authoritative Team consensus, and NDCG at 10 (NDCG@10) to measure overall executable ranking quality. Finally, for the full HiCoMER pipeline, we report ROUGE-L and Decision F1 to evaluate whether improvements in maintained-state retrieval translate into more accurate and executable memory-grounded answers. Detailed metric definitions are provided in Appendix~\ref{sec:appendix_baselines_metrics}.

\subsection{Implementation Details}
\label{sec:impl}

We evaluate HiCoMER on the ALI and PROTAC datasets under two backbone settings. In the strong-backbone setting, both the Hierarchical Memory Conflict Updater and Memory-Grounded Answer Generator use Llama-3.1-8B. In the lightweight setting, both use Qwen2.5-3B-Instruct. The Validity-Aware Memory Retriever is shared across settings. All experiments use the same grouped memory streams, query interface, and downstream prompt templates to ensure fair comparison.

Training proceeds sequentially. First, the updater is trained on structured maintenance instances to learn hierarchical memory conflict resolution. The trained updater generates maintained memory states for the training split, which are then used to train the Validity-Aware Memory Retriever with a pairwise ranking objective. At inference, maintained memories are retrieved and reranked for executability, and the top memories are passed to the answer generator.

\begin{table*}[htbp]
\caption{End-to-end evaluation of the full HiCoMER pipeline on the ALI dataset using Llama-3.1-8B as the backbone for the Hierarchical Memory Conflict Updater and the Memory-Grounded Answer Generator. Retrieval metrics ORR@5, CRR@5, and NDCG@10 measure maintained-state retrieval quality, while answer-level metrics ROUGE-L and Decision F1 assess how well the pipeline converts maintained and retrieved memories into final memory-grounded responses.}
\label{tab:main_results_ali_llama}
\centering
\resizebox{\textwidth}{!}{
\begin{tabular}{lccccc}
\toprule
\textbf{Method} & \textbf{ORR@5 ($\downarrow$)} & \textbf{CRR@5 ($\uparrow$)} & \textbf{Decision F1 ($\uparrow$)} & \textbf{ROUGE-L ($\uparrow$)} & \textbf{NDCG@10 ($\uparrow$)} \\
\midrule
\multicolumn{6}{l}{\textit{Standard \& Semantic Retrieval}} \\
Flat RAG                & 45.83 & 51.94 & 41.67 & 35.38 & 42.76 \\
Hybrid RAG              & 41.92 & 55.68 & 44.27 & 37.14 & 45.31 \\
Rerank RAG              & 37.57 & 61.83 & 48.88 & 40.47 & 51.86 \\
\midrule
\multicolumn{6}{l}{\textit{Multi-Agent \& Hierarchical Memory}} \\
G-Memory                & 29.74 & 68.91 & 53.79 & 44.82 & 57.63 \\
Agentic Memory (Park)   & 32.81 & 64.66 & 52.14 & 43.58 & 54.77 \\
\midrule
\multicolumn{6}{l}{\textit{Reflection \& Production Systems}} \\
Self-RAG                & 35.12 & 62.57 & 55.23 & 45.97 & 53.64 \\
MemGPT-style Window     & 23.79 & 49.08 & 46.97 & 41.73 & 48.88 \\
Mem0                    & 28.63 & 70.84 & 57.36 & 47.48 & 59.27 \\
\midrule
\textbf{HiCoMER (Llama-3.1-8B)} & \textbf{14.18} & \textbf{84.87} & \textbf{64.61} & \textbf{52.93} & \textbf{68.74} \\
\bottomrule
\end{tabular}}
\end{table*}
\subsection{Results}
\label{sec:results}

We first present the main results on the ALI dataset, our primary in-domain evaluation setting. To assess effectiveness under a stronger backbone, we evaluate HiCoMER with Llama-3.1-8B for both the Hierarchical Memory Conflict Updater and the Memory-Grounded Answer Generator. We also evaluate a lighter deployment-oriented setting using Qwen2.5-3B-Instruct to test the robustness of the framework under constrained compute and memory budgets.

Tables~\ref{tab:main_results_ali_llama} and~\ref{tab:main_results_ali_qwen} report the end-to-end performance of HiCoMER. Retrieval-oriented metrics ORR@5, CRR@5, and NDCG@10 primarily measure maintained-state retrieval quality, capturing the joint effect of the Hierarchical Memory Conflict Updater and the Validity-Aware Memory Retriever. Answer-level metrics ROUGE-L and Decision F1 further evaluate how improvements in retrieval translate into higher-quality memory-grounded responses. Using Qwen2.5-3B-Instruct, HiCoMER maintains strong performance, demonstrating that the maintenance-and-retrieval design is robust to smaller backbone models. Gains are observed in both maintained-state retrieval and final response quality.

Overall, HiCoMER substantially reduces outdated or non-validity-aware retrievals while better preserving authoritative Team consensus. This leads to improved executable ranking and stronger downstream responses. The consistent gains across both Llama-3.1-8B and Qwen2.5-3B-Instruct suggest that improvements stem from the framework design rather than backbone scale. These results highlight that semantic relevance alone is insufficient in hierarchical collaborative memory settings: neither stronger semantic matching, recency heuristics, nor generation-time reflection can fully replace explicit memory maintenance and validity-aware retrieval. Further component-level evaluation of the Hierarchical Memory Conflict Updater on held-out structured maintenance instances is provided in Appendix~\ref{sec:appendix_updater_eval}.

Additional results on the PROTAC dataset and extended per-baseline analyses are provided in Appendix~\ref{sec:appendix_protac_results}.

\begin{table*}[t]
\caption{End-to-end evaluation of HiCoMER on the ALI dataset using Qwen2.5-3B-Instruct for both the Hierarchical Memory Conflict Updater and the Memory-Grounded Answer Generator. Retrieval metrics ORR@5, CRR@5, and NDCG@10 assess maintained-state ranking, while answer-level metrics ROUGE-L and Decision F1 measure the quality of memory-grounded responses under a smaller backbone.}
\label{tab:main_results_ali_qwen}
\centering
\resizebox{\textwidth}{!}{
\begin{tabular}{lccccc}
\toprule
\textbf{Method} & \textbf{ORR@5 ($\downarrow$)} & \textbf{CRR@5 ($\uparrow$)} & \textbf{Decision F1 ($\uparrow$)} & \textbf{ROUGE-L ($\uparrow$)} & \textbf{NDCG@10 ($\uparrow$)} \\
\midrule
\multicolumn{6}{l}{\textit{Standard \& Semantic Retrieval}} \\
Flat RAG                & 46.57 & 50.86 & 39.43 & 33.57 & 41.93 \\
Hybrid RAG              & 43.04 & 54.88 & 42.16 & 35.61 & 44.36 \\
Rerank RAG              & 38.76 & 60.43 & 46.48 & 38.92 & 50.37 \\
\midrule
\multicolumn{6}{l}{\textit{Multi-Agent \& Hierarchical Memory}} \\
G-Memory                & 31.03 & 67.12 & 50.84 & 42.69 & 55.87 \\
Agentic Memory (Park)   & 34.08 & 63.04 & 49.58 & 41.76 & 53.41 \\
\midrule
\multicolumn{6}{l}{\textit{Reflection \& Production Systems}} \\
Self-RAG                & 36.77 & 60.93 & 52.09 & 43.46 & 51.94 \\
MemGPT-style Window     & 24.46 & 47.18 & 44.19 & 39.54 & 46.72 \\
Mem0                    & 30.18 & 68.37 & 53.88 & 44.87 & 57.12 \\
\midrule
\textbf{HiCoMER (Qwen2.5-3B-Instruct)} & \textbf{16.07} & \textbf{82.34} & \textbf{60.28} & \textbf{49.81} & \textbf{66.08} \\
\bottomrule
\end{tabular}}
\end{table*}

\subsection{Ablation Study}
\label{sec:ablation}

To quantify the contribution of each major module in HiCoMER, we conduct a full-pipeline ablation study on the ALI dataset. We remove each of the three core modules individually: the Hierarchical Memory Conflict Updater, the Validity-Aware Memory Retriever, and the Memory-Grounded Answer Generator. Table~\ref{tab:ablation} reports both retrieval-oriented metrics and answer-level metrics to assess how degradations in upstream modules propagate to final response quality.

\begin{table*}[t]
\caption{Ablation study on the ALI dataset. ORR@5, CRR@5, and NDCG@10 measure maintained-state retrieval quality, while ROUGE-L and Decision F1 assess memory-grounded response quality. Each row shows the effect of removing a single module from the full HiCoMER pipeline.}
\label{tab:ablation}
\centering
\resizebox{\textwidth}{!}{
\begin{tabular}{lccccc}
\toprule
Variant & ORR@5 & CRR@5 & NDCG@10 & ROUGE-L & Decision F1 \\
\midrule
Full HiCoMER (Llama-3.1-8B) & 14.18 & 84.87 & 68.74 & 52.93 & 64.61 \\
\quad w/o Hierarchical Memory Conflict Updater & 30.42 & 71.63 & 58.91 & 46.86 & 55.94 \\
\quad w/o Validity-Aware Memory Retriever & 29.16 & 70.31 & 57.12 & 45.97 & 54.83 \\
\quad w/o Memory-Grounded Answer Generator & 14.18 & 84.87 & 68.74 & 47.62 & 56.48 \\
\bottomrule
\end{tabular}}
\end{table*}

Removing either the updater or the validity-aware retriever leads to substantial drops in both maintained-state retrieval metrics and final response quality, confirming that these upstream modules are critical for end-to-end performance. Removing the answer generator also degrades final response metrics, highlighting its role in converting retrieved memories into high-quality answers. Additional ablation analyses on the PROTAC dataset are provided in Appendix~\ref{sec:appendix_ablation_protac}.

\subsection{Human Evaluation}
\label{sec:human_eval}

We conduct a human evaluation to assess the practical performance of our proposed method. Seventeen participants with research experience, including Ph.D. students, master's students, and junior research assistants, evaluated 10 report-writing tasks each, yielding a total of 170 evaluation instances. For each task, participants evaluated outputs generated under the same scenario by HiCoMER and Mem0. To reduce potential evaluation bias, the outputs were presented in randomized order, and participants were not informed which method produced each output.

We evaluate the outputs of HiCoMER and Mem0 in terms of two aspects, Memory Precision and Intervention Usefulness. Memory Precision measures the relevance and accuracy of retrieved context, while Intervention Usefulness measures the actionability of the generated suggestions. Participants rated each aspect on a 5-point scale, with a score of 5 indicating the highest performance. 

\begin{table}[t]
\centering
\caption{Human evaluation results on simulated system utility, based on 170 evaluation instances. Memory Precision measures the relevance and accuracy of retrieved context, and Intervention Usefulness measures the actionability of the generated suggestions.}
\label{tab:human_scores}
\resizebox{\columnwidth}{!}{
\begin{tabular}{lcc}
\toprule
Method & Memory Precision & Intervention Usefulness \\
\midrule
Mem0 & 3.73 & 3.68 \\
HiCoMER & 4.11 & 4.04 \\
\bottomrule
\end{tabular}
}
\end{table}

The human evaluation results are reported in Table~\ref{tab:human_scores}. HiCoMER received higher ratings than Mem0 on both aspects of Memory Precision and Intervention Usefulness, suggesting that it provides more accurate memory support and more actionable suggestions in realistic scientific collaboration scenarios.

\section{Conclusion}
We present HiCoMER, a hierarchical collaborative memory framework for long-horizon LLM agents. HiCoMER models team-level and individual-level information, maintains memory consistency with a trainable hierarchical conflict updater, retrieves currently valid memories via a validity-aware scorer, and generates final answers grounded in the maintained evidence. Experimental results on the datasets constructed in this work show that HiCoMER substantially reduces outdated retrieval while preserving valid memory states, improving downstream tasks and yielding higher-quality, memory-grounded answers and positive human evaluation outcomes. Our results further suggest that effective long-term memory management requires not only storing and retrieving historical information, but also identifying which memories remain valid as the collaborative context evolves. Future work includes richer conflict structures, multi-level supersession, and robustness under noisier or asynchronous memory streams.

\section*{Limitations}

While HiCoMER demonstrates consistent improvements in hierarchical memory maintenance and validity-aware retrieval, a few limitations should be noted. First, noisy, incomplete, or misaligned Team or Individual records could affect maintenance accuracy and downstream answer quality. Finally, our current evaluation focuses on biomedical collaboration scenarios. Future work will extend testing to additional domains. 

\section*{Ethical Statement}
HiCoMER is designed for research and simulation of collaborative memory management in LLM agents and does not directly interact with human subjects or sensitive patient data in deployed settings. All datasets used in this work are derived from publicly available scientific publications and are reverse-simulated to avoid exposing identifiable human data. 

For the human evaluation, participants were recruited voluntarily from research-experienced individuals and performed scenario-based tasks on simulated data only. No private or sensitive information was used, and all participant data were anonymized. Participants provided informed consent and were debriefed after the evaluation.

\bibliography{custom}

@article{boiko2023autonomous,
  title={Autonomous chemical research with large language models},
  author={Boiko, Daniil A and MacKnight, Robert and Kline, Ben and Gomes, Gabe},
  journal={Nature},
  volume={624},
  number={7992},
  pages={570--578},
  year={2023},
  publisher={Nature Publishing Group UK London}
}

@inproceedings{lewis2020retrieval,
  author       = {Patrick Lewis and
                  Ethan Perez and
                  Aleksandra Piktus and
                  Fabio Petroni and
                  Vladimir Karpukhin and
                  Naman Goyal and
                  Heinrich K{\"{u}}ttler and
                  Mike Lewis and
                  Wen{-}tau Yih and
                  Tim Rockt{\"{a}}schel and
                  Sebastian Riedel and
                  Douwe Kiela},
  title        = {Retrieval-Augmented Generation for Knowledge-Intensive {NLP} Tasks},
  booktitle    = {Advances in Neural Information Processing Systems 33: Annual Conference
                  on Neural Information Processing Systems 2020, NeurIPS 2020},
  year         = {2020}
}

@article{packer2023memgpt,
  title={Memgpt: Towards llms as operating systems},
  author={Packer, Charles and Wooders, Sarah and Lin, Kevin and Fang, Vivian and Patil, Shishir G and Stoica, Ion and Gonzalez, Joseph E},
  journal={arXiv preprint arXiv:2310.08560},
  year={2023}
}

@inproceedings{park2023generativeagents,
  author       = {Joon Sung Park and
                  Joseph C. O'Brien and
                  Carrie Jun Cai and
                  Meredith Ringel Morris and
                  Percy Liang and
                  Michael S. Bernstein},
  title        = {Generative Agents: Interactive Simulacra of Human Behavior},
  booktitle    = {Proceedings of the 36th Annual {ACM} Symposium on User Interface Software and Technology},
  pages        = {1--22},
  year         = {2023}
}

@article{gao2024biomedical,
  title={Empowering biomedical discovery with AI agents},
  author={Gao, Shanghua and Fang, Ada and Huang, Yepeng and Giunchiglia, Valentina and Noori, Ayush and Schwarz, Jonathan Richard and Ektefaie, Yasha and Kondic, Jovana and Zitnik, Marinka},
  journal={Cell},
  volume={187},
  number={22},
  pages={6125--6151},
  year={2024}
}

@article{chhikara2025mem0,
  title={Mem0: Building production-ready ai agents with scalable long-term memory},
  author={Chhikara, Prateek and Khant, Dev and Aryan, Saket and Singh, Taranjeet and Yadav, Deshraj},
  journal={arXiv preprint arXiv:2504.19413},
  year={2025}
}

@inproceedings{zhang2025gmemory,
  author       = {Guibin Zhang and
                  Muxin Fu and
                  Kun Wang and
                  Frank Wan and
                  Miao Yu and
                  Shuicheng Yan},
  title        = {G-Memory: Tracing Hierarchical Memory for Multi-Agent Systems},
  booktitle    = {Advances in Neural Information Processing Systems 38: Annual Conference
                  on Neural Information Processing Systems 2025, NeurIPS 2025},
  year         = {2025}
}

@inproceedings{karpukhin2020dpr,
  title={Dense passage retrieval for open-domain question answering},
  author={Karpukhin, Vladimir and Oguz, Barlas and Min, Sewon and Lewis, Patrick and Wu, Ledell and Edunov, Sergey and Chen, Danqi and Yih, Wen-tau},
  booktitle={Proceedings of the 2020 conference on empirical methods in natural language processing (EMNLP)},
  pages={6769--6781},
  year={2020}
}

@inproceedings{guu2020realm,
  title={Retrieval augmented language model pre-training},
  author={Guu, Kelvin and Lee, Kenton and Tung, Zora and Pasupat, Panupong and Chang, Mingwei},
  booktitle={International conference on machine learning},
  pages={3929--3938},
  year={2020}
}

@inproceedings{borgeaud2022retro,
  author       = {Sebastian Borgeaud and
                  Arthur Mensch and
                  Jordan Hoffmann and
                  Trevor Cai and
                  Eliza Rutherford and
                  Katie Millican and
                  George van den Driessche and
                  Jean{-}Baptiste Lespiau and
                  Bogdan Damoc and
                  Aidan Clark and
                  Diego de Las Casas and
                  Aurelia Guy and
                  Jacob Menick and
                  Roman Ring and
                  Tom Hennigan and
                  Saffron Huang and
                  Loren Maggiore and
                  Chris Jones and
                  Albin Cassirer and
                  Andy Brock and
                  Michela Paganini and
                  Geoffrey Irving and
                  Oriol Vinyals and
                  Simon Osindero and
                  Karen Simonyan and
                  Jack W. Rae and
                  Erich Elsen and
                  Laurent Sifre},
  title        = {Improving Language Models by Retrieving from Trillions of Tokens},
  booktitle    = {International Conference on Machine Learning},
  pages        = {2206--2240},
  year         = {2022}
}

@inproceedings{izacard2021fid,
  title={Leveraging passage retrieval with generative models for open domain question answering},
  author={Izacard, Gautier and Grave, Edouard},
  booktitle={Proceedings of the 16th conference of the european chapter of the association for computational linguistics: main volume},
  pages={874--880},
  year={2021}
}

@article{izacard2023atlas,
  title={Atlas: Few-shot learning with retrieval augmented language models},
  author={Izacard, Gautier and Lewis, Patrick and Lomeli, Maria and Hosseini, Lucas and Petroni, Fabio and Schick, Timo and Dwivedi-Yu, Jane and Joulin, Armand and Riedel, Sebastian and Grave, Edouard},
  journal={Journal of Machine Learning Research},
  volume={24},
  number={251},
  pages={1--43},
  year={2023}
}

@inproceedings{khattab2020colbert,
  title={Colbert: Efficient and effective passage search via contextualized late interaction over bert},
  author={Khattab, Omar and Zaharia, Matei},
  booktitle={Proceedings of the 43rd International ACM SIGIR conference on research and development in Information Retrieval},
  pages={39--48},
  year={2020}
}

@inproceedings{formal2021splade,
  title={SPLADE: Sparse lexical and expansion model for first stage ranking},
  author={Formal, Thibault and Piwowarski, Benjamin and Clinchant, St{\'e}phane},
  booktitle={Proceedings of the 44th international ACM SIGIR conference on research and development in information retrieval},
  pages={2288--2292},
  year={2021}
}

@inproceedings{gao2023hyde,
  title={Precise zero-shot dense retrieval without relevance labels},
  author={Gao, Luyu and Ma, Xueguang and Lin, Jimmy and Callan, Jamie},
  booktitle={Proceedings of the 61st Annual Meeting of the Association for Computational Linguistics (Volume 1: Long Papers)},
  pages={1762--1777},
  year={2023}
}

@inproceedings{bowman2015snli,
  title={A large annotated corpus for learning natural language inference},
  author={Bowman, Samuel R and Angeli, Gabor and Potts, Christopher and Manning, Christopher D},
  booktitle={Proceedings of the 2015 conference on empirical methods in natural language processing},
  pages={632--642},
  year={2015}
}

@inproceedings{williams2018mnli,
  title={A broad-coverage challenge corpus for sentence understanding through inference},
  author={Williams, Adina and Nangia, Nikita and Bowman, Samuel R},
  booktitle={Proceedings of the 2018 conference of the North American chapter of the association for computational linguistics: human language technologies, volume 1 (long papers)},
  pages={1112--1122},
  year={2018}
}

@inproceedings{nie2020anli,
  title={Adversarial NLI: A new benchmark for natural language understanding},
  author={Nie, Yixin and Williams, Adina and Dinan, Emily and Bansal, Mohit and Weston, Jason and Kiela, Douwe},
  booktitle={Proceedings of the 58th annual meeting of the association for computational linguistics},
  pages={4885--4901},
  year={2020}
}

@inproceedings{thorne2018fever,
  title={FEVER: a large-scale dataset for fact extraction and VERification},
  author={Thorne, James and Vlachos, Andreas and Christodoulopoulos, Christos and Mittal, Arpit},
  booktitle={Proceedings of the 2018 Conference of the North American Chapter of the Association for Computational Linguistics: Human Language Technologies, Volume 1 (Long Papers)},
  pages={809--819},
  year={2018}
}

@inproceedings{kryscinski2020factcc,
  title={Evaluating the factual consistency of abstractive text summarization},
  author={Kry{\'s}ci{\'n}ski, Wojciech and McCann, Bryan and Xiong, Caiming and Socher, Richard},
  booktitle={Proceedings of the 2020 conference on empirical methods in natural language processing (EMNLP)},
  pages={9332--9346},
  year={2020}
}

@inproceedings{wang2020qags,
  title={Asking and answering questions to evaluate the factual consistency of summaries},
  author={Wang, Alex and Cho, Kyunghyun and Lewis, Mike},
  booktitle={Proceedings of the 58th annual meeting of the association for computational linguistics},
  pages={5008--5020},
  year={2020}
}

@inproceedings{lin2022truthfulqa,
  title={Truthfulqa: Measuring how models mimic human falsehoods},
  author={Lin, Stephanie and Hilton, Jacob and Evans, Owain},
  booktitle={Proceedings of the 60th annual meeting of the association for computational linguistics (volume 1: long papers)},
  pages={3214--3252},
  year={2022}
}

@inproceedings{honovich2022true,
  title={TRUE: Re-evaluating factual consistency evaluation},
  author={Honovich, Or and Aharoni, Roee and Herzig, Jonathan and Taitelbaum, Hagai and Kukliansy, Doron and Cohen, Vered and Scialom, Thomas and Szpektor, Idan and Hassidim, Avinatan and Matias, Yossi},
  booktitle={Proceedings of the 2022 Conference of the North American Chapter of the Association for Computational Linguistics: Human Language Technologies},
  pages={3905--3920},
  year={2022}
}

@inproceedings{manakul2023selfcheckgpt,
  title={Selfcheckgpt: Zero-resource black-box hallucination detection for generative large language models},
  author={Manakul, Potsawee and Liusie, Adian and Gales, Mark},
  booktitle={Proceedings of the 2023 conference on empirical methods in natural language processing},
  pages={9004--9017},
  year={2023}
}

@inproceedings{gao2023rarr,
  title={Rarr: Researching and revising what language models say, using language models},
  author={Gao, Luyu and Dai, Zhuyun and Pasupat, Panupong and Chen, Anthony and Chaganty, Arun Tejasvi and Fan, Yicheng and Zhao, Vincent and Lao, Ni and Lee, Hongrae and Juan, Da-Cheng and Guu, Kelvin},
  booktitle={Proceedings of the 61st Annual Meeting of the Association for Computational Linguistics (Volume 1: Long Papers)},
  pages={16477--16508},
  year={2023}
}

@inproceedings{min2023factscore,
  title={FActScore: Fine-grained atomic evaluation of factual precision in long form text generation},
  author={Min, Sewon and Krishna, Kalpesh and Lyu, Xinxi and Lewis, Mike and Yih, Wen-tau and Koh, Pang and Iyyer, Mohit and Zettlemoyer, Luke and Hajishirzi, Hannaneh},
  booktitle={Proceedings of the 2023 conference on empirical methods in natural language processing},
  pages={12076--12100},
  year={2023}
}

@article{shuster2022blenderbot,
  title={Blenderbot 3: a deployed conversational agent that continually learns to responsibly engage},
  author={Shuster, Kurt and Xu, Jing and Komeili, Mojtaba and Ju, Da and Smith, Eric Michael and Roller, Stephen and Ung, Megan and Chen, Moya and Arora, Kushal and Lane, Joshua and Behrooz, Morteza and Ngan, William and Poff, Spencer and Goyal, Naman and Szlam, Arthur and Boureau, Y-Lan and Kambadur, Melanie and Weston, Jason},
  journal={arXiv preprint arXiv:2208.03188},
  year={2022}
}

@inproceedings{asai2024self,
  title={Self-rag: Learning to retrieve, generate, and critique through self-reflection},
  author={Asai, Akari and Wu, Zeqiu and Wang, Yizhong and Sil, Avi and Hajishirzi, Hannaneh},
  booktitle={International conference on learning representations},
  volume={2024},
  pages={9112--9141},
  year={2024}
}

@article{shao2024deepseekmath,
  title={DeepSeekMath: Pushing the Limits of Mathematical Reasoning in Open Language Models},
  author={Shao, Zhihong and Wang, Peiyi and Zhu, Qihao and Xu, Runxin and Song, Junxiao and Bi, Xiao and Zhang, Haowei and Zhang, Mingchuan and Li, Y. K. and Wu, Y. and Guo, Daya},
  journal={arXiv preprint arXiv:2402.03300},
  year={2024}
}

\appendix

\section{Detailed Conflict Resolution in Answer Generation}
\label{app:conflict_resolution}

Given two retrieved memories $m_i$ and $m_j$, HiCoMER checks if their maintained contents are contradictory:
\begin{equation}
\mathrm{Contradict}(x_{m_i}^{\mathrm{maint}},x_{m_j}^{\mathrm{maint}})=1.
\end{equation}

If no contradiction is detected, both memories are retained. If a contradiction exists, the memory with the higher priority is retained:
\begin{equation}
\begin{split}
m_i \succ m_j \quad \text{if} \quad & 
\mathrm{Contradict}(x_{m_i}^{\mathrm{maint}},x_{m_j}^{\mathrm{maint}})=1 \\
& \land \ P(m_i) > P(m_j).
\end{split}
\end{equation}

The final evidence set for answer generation is then defined as
\begin{equation}
\begin{split}
\mathcal{C}^{*} = \{ m \in R_t(q) \ \mid \ & 
\nexists\, m' \in R_t(q) \ \text{such that} \\
& m' \succ m \}.
\end{split}
\end{equation}

All else being equal, this procedure yields the following preference order in common cases:
\begin{itemize}
    \item Team memories are preferred over Individual memories when their maintained contents conflict.
    \item Among memories of the same type, those that remain valid under the maintained state are preferred over invalid or outdated ones.
    \item If both memories have the same type and validity, the memory with the more recent maintenance time is preferred.
\end{itemize}

After conflict resolution, each memory in $\mathcal{C}^{*}$ is serialized with its source type, maintenance time, and maintained content, and the resulting structured evidence set is passed to the instruction-following LLM for final answer generation.

\section{ALI and PROTAC Dataset Construction Details}
\label{sec:appendix_dataset}

This appendix provides full details on the construction of the ALI and PROTAC datasets. For each domain, we start with a candidate pool of 30 PubMed papers. After paper-level eligibility screening, 27 ALI papers and 26 PROTAC papers are retained. Each retained paper is reverse-simulated into a 12-week collaborative lifecycle, divided into event windows spanning 48--72 hours. Each event window corresponds to a grouped memory unit containing Team memories, which encode collective decisions, protocol-level constraints, and authoritative consensus, and Individual memories, which capture local execution traces, observations, and intermediate progress.

Hierarchical conflicts are explicitly injected, including same-time conflicts within a single event window and cross-time conflicts spanning multiple windows. ALI contains 317 injected conflicts (189 same-time, 128 cross-time), and PROTAC contains 304 injected conflicts (181 same-time, 123 cross-time). To reduce document-level leakage, train/validation/test splits are organized by source paper: ALI uses 19/3/5 and PROTAC uses 18/3/5. Both datasets support six query types: fact lookup, procedure validation, conflict diagnosis, risk warning, next-step decision, and protocol consistency checking. GPT-5 is used for reverse simulation, query-answer generation, and evidence-trace annotation. Across valid groups, each contains on average 2.1 Team memories and 7.8 Individual memories, totaling 9.9 documents per group.

\section{Illustrative examples}
\label{sec:appendix_illustrative_examples}
Illustrative examples for the ALI and PROTAC datasets are provided in Table~\ref{tab:illustrative_examples}.

\begin{table*}[t]
\caption{Illustrative examples.}
\label{tab:illustrative_examples}
\centering
\resizebox{\textwidth}{!}{
\begin{tabular}{p{2cm}p{4.2cm}p{4.6cm}p{4.8cm}}
\toprule
\textbf{Dataset} & \textbf{Team Memory} & \textbf{Conflicting Individual/ Historical Memory} & \textbf{Query} \\
\midrule
ALI &
Week 4 team decision: discontinue Compound X due to elevated toxicity in the pilot cohort. &
Week 4 lab log: prepared Compound X for follow-up efficacy testing; Week 2 protocol note: expand Compound X testing if preliminary efficacy remains stable. &
Should the team proceed with the planned Compound X follow-up experiment? \\
\midrule
PROTAC &
Week 6 platform update: replace the original degradation assay with the N-end-rule-based validation workflow for downstream screening. &
Week 6 experiment log: continued using the previous degradation readout; Week 4 design note: prioritize the earlier target-screening route. &
Is the current validation result directly usable under the latest platform protocol? \\
\bottomrule
\end{tabular}}
\end{table*}

\begin{table*}[htbp]
\caption{Component-level evaluation of the Hierarchical Memory Conflict Updater on held-out structured maintenance instances from ALI. Conflict F1 measures the accuracy of predicting conflict relations, while Maintenance Accuracy measures the correctness of structured memory updates.}
\label{tab:appendix_updater_eval}
\centering
\begin{tabular}{lcc}
\toprule
Model & Conflict F1 & Maintenance Accuracy \\
\midrule
Rule-based Updater & 46.13 & 53.38 \\
Qwen2.5-3B Zero-shot & 57.84 & 60.57 \\
Qwen2.5-3B with SFT & 77.18 & 81.47 \\
Qwen2.5-3B with SFT and GRPO & 83.57 & 87.31 \\
\midrule
Llama-3.1-8B Zero-shot & 61.68 & 64.93 \\
Llama-3.1-8B with SFT & 81.76 & 85.36 \\
Llama-3.1-8B with SFT and GRPO & 87.88 & 91.17 \\
\bottomrule
\end{tabular}
\end{table*}
\section{Baseline Definitions and Metric Details}
\label{sec:appendix_baselines_metrics}

To evaluate HiCoMER's handling of hierarchical memory conflicts, we compare it against several baseline categories. All baselines are evaluated on the same grouped memory stream and query interface to ensure differences reflect memory maintenance and retrieval strategies rather than input format or generation capacity.

\begin{itemize}
    \item \textbf{Standard retrieval and semantic matching:} 
    Flat RAG ranks candidates by dense semantic similarity. Hybrid RAG combines dense and sparse (BM25) retrieval. Rerank RAG applies a cross-encoder to rerank Hybrid RAG candidates. These provide strong semantic matching baselines.

    \item \textbf{Recency-based memory control:} 
    MemGPT-style Window retains only the most recent memories within a fixed active window, simulating recency-focused memory management.

    \item \textbf{Agentic and hierarchical memory systems:} 
    Agentic Memory scores candidates by relevance, recency, and static importance. G-Memory organizes memories hierarchically and performs structured retrieval over the shared memory pool.

    \item \textbf{Reflection and production memory frameworks:} 
    Self-RAG applies generation-time reflection to assess whether retrieved memories remain valid. Mem0 is a production memory infrastructure adapted to the shared-memory benchmark.

    \item \textbf{Internal HiCoMER variants:} 
    HiCoMER without validity-aware retrieval retains the maintenance module but ranks memories solely by semantic similarity.

    \item \textbf{Updater-only baselines:} 
    Rule-based Updater applies deterministic revise rules. Qwen2.5-3B Zero-shot uses direct structured prompting without training. Qwen2.5-3B + SFT applies supervised fine-tuning. Qwen2.5-3B + SFT + GRPO is the full trainable updater. Llama-3.1-8B Zero-shot/SFT/SFT+GRPO provide larger-backbone variants for robustness evaluation.
\end{itemize}

Metrics are defined to assess multiple dimensions of performance:

\begin{itemize}
    \item \textbf{Updater-specific metrics:} Conflict F1 measures accuracy in predicting conflict relations. Maintenance Accuracy measures exact-match correctness of relation and action labels.

    \item \textbf{Retrieval-oriented metrics:} ORR@K (Outdated Retrieval Rate) is the fraction of retrieved memories labeled outdated or non-executable. CRR@K (Consensus Retention Rate) is the recall of authoritative Team memories. NDCG@10 evaluates ranking quality with higher gain assigned to valid supporting memories.

    \item \textbf{Answer-level full-pipeline metrics:} ROUGE-L measures textual overlap between generated and gold answers. Decision F1 evaluates correctness on decision-oriented queries, including risk warnings, next-step decisions, and protocol consistency checks.
\end{itemize}

Conflict F1 and Maintenance Accuracy primarily assess the updater, ORR@5, CRR@5, and NDCG@10 reflect the combined effect of the updater and the validity-aware memory retriever, and ROUGE-L and Decision F1 evaluate the end-to-end quality of memory-grounded answer generation.

\section{Updater Evaluation}
\label{sec:appendix_updater_eval}

We evaluate the Hierarchical Memory Conflict Updater on held-out structured maintenance instances derived from the ALI dataset. This component-level evaluation measures whether the updater can correctly predict conflict relations and structured maintenance actions.

The results show that the trained updater substantially outperforms rule-based maintenance, indicating that hierarchical memory updating cannot be reduced to standard contradiction detection alone. Reward-based GRPO optimization further improves structured executable-state maintenance, yielding consistent gains over zero-shot and SFT-only settings.

\begin{table*}[htbp]
\caption{End-to-end evaluation of the full HiCoMER pipeline on the PROTAC dataset using Llama-3.1-8B as the backbone for the Hierarchical Memory Conflict Updater and the Memory-Grounded Answer Generator. Retrieval metrics ORR@5, CRR@5, and NDCG@10 measure maintained-state retrieval quality, while answer-level metrics ROUGE-L and Decision F1 assess how well the pipeline converts maintained and retrieved memories into final memory-grounded responses.}
\label{tab:main_results_protac_llama}
\centering
\resizebox{\textwidth}{!}{
\begin{tabular}{lccccc}
\toprule
\textbf{Method} & \textbf{ORR@5 ($\downarrow$)} & \textbf{CRR@5 ($\uparrow$)} & \textbf{Decision F1 ($\uparrow$)} & \textbf{ROUGE-L ($\uparrow$)} & \textbf{NDCG@10 ($\uparrow$)} \\
\midrule
\multicolumn{6}{l}{\textit{Standard \& Semantic Retrieval}} \\
Flat RAG                & 48.12 & 48.73 & 38.66 & 33.81 & 40.68 \\
Hybrid RAG              & 44.79 & 52.14 & 41.08 & 35.23 & 43.37 \\
Rerank RAG              & 40.66 & 57.58 & 45.42 & 38.17 & 49.09 \\
\midrule
\multicolumn{6}{l}{\textit{Multi-Agent \& Hierarchical Memory}} \\
G-Memory                & 31.84 & 64.77 & 50.31 & 42.38 & 54.34 \\
Agentic Memory (Park)   & 34.87 & 61.92 & 48.79 & 41.14 & 51.83 \\
\midrule
\multicolumn{6}{l}{\textit{Reflection \& Production Systems}} \\
Self-RAG                & 37.88 & 60.71 & 51.63 & 43.76 & 50.42 \\
MemGPT-style Window     & 25.47 & 45.63 & 43.18 & 38.86 & 45.74 \\
Mem0                    & 30.69 & 66.53 & 53.07 & 45.12 & 55.48 \\
\midrule
\textbf{HiCoMER (Llama-3.1-8B)} & \textbf{15.37} & \textbf{82.74} & \textbf{62.79} & \textbf{51.43} & \textbf{66.81} \\
\bottomrule
\end{tabular}}
\end{table*}
\section{Results on PROTAC Dataset}
\label{sec:appendix_protac_results}

\begin{table*}[htbp]
\caption{End-to-end evaluation of HiCoMER on the PROTAC dataset using Qwen2.5-3B-Instruct for both the Hierarchical Memory Conflict Updater and the Memory-Grounded Answer Generator. Retrieval metrics ORR@5, CRR@5, and NDCG@10 assess maintained-state ranking, while answer-level metrics ROUGE-L and Decision F1 measure the quality of memory-grounded responses under a smaller backbone.}
\label{tab:main_results_protac_qwen}
\centering
\resizebox{\textwidth}{!}{
\begin{tabular}{lccccc}
\toprule
\textbf{Method} & \textbf{ORR@5 ($\downarrow$)} & \textbf{CRR@5 ($\uparrow$)} & \textbf{Decision F1 ($\uparrow$)} & \textbf{ROUGE-L ($\uparrow$)} & \textbf{NDCG@10 ($\uparrow$)} \\
\midrule
\multicolumn{6}{l}{\textit{Standard \& Semantic Retrieval}} \\
Flat RAG                & 48.97 & 47.63 & 36.88 & 31.94 & 39.47 \\
Hybrid RAG              & 45.74 & 50.82 & 39.42 & 33.76 & 42.03 \\
Rerank RAG              & 41.48 & 56.07 & 43.61 & 36.87 & 47.58 \\
\midrule
\multicolumn{6}{l}{\textit{Multi-Agent \& Hierarchical Memory}} \\
G-Memory                & 33.02 & 63.09 & 47.86 & 40.58 & 52.46 \\
Agentic Memory (Park)   & 35.93 & 59.97 & 46.82 & 39.53 & 50.63 \\
\midrule
\multicolumn{6}{l}{\textit{Reflection \& Production Systems}} \\
Self-RAG                & 39.06 & 58.88 & 49.18 & 41.74 & 48.66 \\
MemGPT-style Window     & 26.08 & 43.87 & 40.76 & 36.83 & 44.19 \\
Mem0                    & 31.87 & 64.66 & 50.73 & 42.93 & 53.88 \\
\midrule
\textbf{HiCoMER (Qwen2.5-3B-Instruct)} & \textbf{17.58} & \textbf{79.83} & \textbf{58.12} & \textbf{48.17} & \textbf{63.96} \\
\bottomrule
\end{tabular}}
\end{table*}

We present the full main results on the PROTAC dataset in this appendix. As in the main text, the tables report end-to-end evaluations of the complete HiCoMER pipeline. Metrics {ORR@5}, {CRR@5}, and {NDCG@10} primarily capture the combined effect of the {Hierarchical Memory Conflict Updater} and {Validity-Aware Memory Retriever}, while {ROUGE-L} and {Decision F1} assess the final response quality produced by the {Memory-Grounded Answer Generator}.

\begin{table*}[htbp]
\caption{Ablation study on the PROTAC dataset. ORR@5, CRR@5, and NDCG@10 measure maintained-state retrieval quality, while ROUGE-L and Decision F1 assess memory-grounded response quality. Each row shows the effect of removing a single module from the full HiCoMER pipeline.}
\label{tab:appendix_ablation_protac}
\centering
\resizebox{\textwidth}{!}{
\begin{tabular}{lccccc}
\toprule
Variant & ORR@5 & CRR@5 & NDCG@10 & ROUGE-L & Decision F1 \\
\midrule
Full HiCoMER (Llama-3.1-8B) & 15.37 & 82.74 & 66.81 & 51.43 & 62.79 \\
\quad w/o Hierarchical Memory Conflict Updater & 31.26 & 70.94 & 57.63 & 45.98 & 54.86 \\
\quad w/o Validity-Aware Memory Retriever & 30.11 & 69.82 & 56.34 & 45.21 & 53.91 \\
\quad w/o Memory-Grounded Answer Generator & 15.37 & 82.74 & 66.81 & 46.58 & 54.97 \\
\bottomrule
\end{tabular}}
\end{table*}

The evaluation on the PROTAC dataset demonstrates that HiCoMER consistently reduces outdated retrieval and preserves valid memory states under complex, evolving workflow constraints. Improvements in maintained-state retrieval translate into higher-quality, memory-grounded responses, reflecting the effective coordination of hierarchical memory maintenance and validity-aware retrieval. Overall, these results indicate that HiCoMER provides a robust and effective framework for managing dynamic collaborative memories in mechanism-driven platform settings.

\section{Ablation Study on the PROTAC Dataset}
\label{sec:appendix_ablation_protac}

To further assess module contributions across datasets, we conduct an ablation study on the PROTAC dataset. Each of the three core HiCoMER modules is removed individually: the Hierarchical Memory Conflict Updater, the Validity-Aware Memory Retriever, and the Memory-Grounded Answer Generator. Table~\ref{tab:appendix_ablation_protac} reports retrieval-oriented and answer-level metrics to evaluate how module removals affect maintained-state retrieval and final response quality.

Removing either the updater or the validity-aware retriever results in significant drops across both retrieval and answer-level metrics, confirming that these upstream modules are essential for end-to-end performance. Removing the answer generator also reduces response quality, demonstrating its role in converting retrieved memories into high-quality answers. These results complement the ALI dataset ablation study and show consistent module contributions across datasets.

\end{document}